\pdfoutput=1

\documentclass[11pt]{article}

\usepackage{acl}

\usepackage{times}
\usepackage{latexsym}
\usepackage[T1]{fontenc}
\usepackage[utf8]{inputenc}
\usepackage{microtype}
\usepackage{inconsolata}

\usepackage{soul}
\usepackage{url}
\usepackage{booktabs}
\usepackage{arydshln}
\usepackage{nicefrac}
\usepackage{multirow}
\usepackage{tabularx}
\usepackage{amsmath}
\usepackage{amssymb}
\usepackage{amsfonts}
\usepackage{mathtools}

\usepackage{xcolor,colortbl}
\usepackage[export]{adjustbox}

\usepackage{makecell}
\usepackage{wrapfig}

\usepackage{tcolorbox}
\tcbuselibrary{listings,breakable}
\tcbset{listing engine=listings,colframe=black,colback=white,size=small}

\usepackage{upquote}
\definecolor{dkgreen}{rgb}{0,0.6,0}
\definecolor{gray}{rgb}{0.5,0.5,0.5}
\definecolor{mauve}{rgb}{0.58,0,0.82}
\usepackage{xspace}

\newcommand{\sense}[1][]{\textsc{Sense}\texttt{#1}\xspace}
\newcommand{\senseq}[1][]{\textsc{Sense}$^{\textrm{Q}}$\texttt{#1}\xspace}

\newcolumntype{x}[1]{>{\centering\arraybackslash\hspace{0pt}}p{#1}}

\title{Synthesizing Text-to-SQL Data from Weak and Strong LLMs}

\author{
\bf
Jiaxi Yang$^{1,2,*}$\footnotemark[3], 
Binyuan Hui$^{3,*}$, 
Min Yang$^{1}$\footnotemark[2],
Jian Yang $^{3}$, 
Junyang Lin$^{3}$, 
Chang Zhou$^{3}$\footnotemark[2] \\
$^1$ Shenzhen Institute of Advanced Technology, Chinese Academy of Sciences \\
$^2$ University of Chinese Academy of Sciences\\
$^3$ Alibaba Group \\
\texttt{\{jx.yang, min.yang\}@siat.ac.cn} \\
\texttt{binyuan.hby@alibaba-inc.com} \\
\url{https://github.com/Yangjiaxi/Sense}
}

\begin{document}

\maketitle

\renewcommand{\thefootnote}{\fnsymbol{footnote}}
\footnotetext{$^{*}$ Equal contribution.}
\footnotetext[3]{Work done during an intern at Alibaba Group.}
\footnotetext[2]{Corresponding authors.}

\begin{abstract}
The capability gap between open-source and closed-source large language models (LLMs) remains challenging in text-to-SQL tasks. 
In this paper, we introduce a synthetic data approach that amalgamates strong data generated by larger, more potent models (strong models) with weak data produced by smaller, less well-aligned models (weak models).
Our approach contributes to the improvement of domain generalization in text-to-SQL models and investigates the potential of weak data supervision through preference learning. Moreover, we utilize the synthetic data approach for instruction tuning on open-source LLMs, yielding \sense, a specialized text-to-SQL model.
The effectiveness of \sense is substantiated by achieving state-of-the-art results on the SPIDER and BIRD benchmarks, thereby mitigating the performance disparity between open-source models and the methods derived from closed-source models.
\end{abstract}

\section{Introduction}

\begin{figure}[!h]
    \centering
    \includegraphics[width=1.0\linewidth]{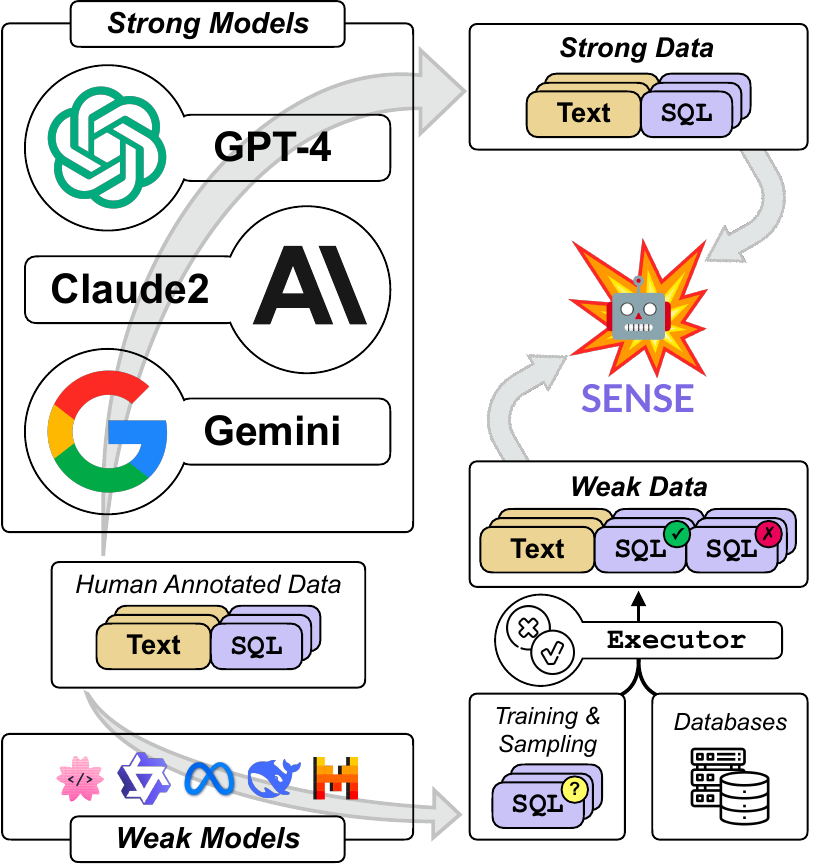}
    \caption{Overview of \sense: Integrating human-annotated data with synthetic data from strong models for domain diversity, and weak models for preference learning, aligning with executors for enhanced text-to-SQL performance.}
    \label{fig:enter-label}
\end{figure}

The ability to convert a natural language question into a structured query language (SQL), i.e., text-to-SQL~\cite{wikiSQL,yu-etal-2018-spider,bird,qin2022survey}, can assist non-experts in interacting with databases using natural language, democratizing data access and analysis. In recent studies~\cite{dailsql,actsql}, notable accomplishments have been observed in powerful closed-source LLMs, exemplified by GPT-4, employing a range of prompting methods~\cite{cot}. However, the adoption of closed-source LLMs introduces concerns pertaining to issues of openness, privacy, and substantial costs.

Recently, the proliferation of numerous open-source LLMs~\cite{llama2,codellama,bai2023qwen} has attracted considerable attention as these models demonstrate comparable capabilities to their closed-source counterparts across a broad spectrum of natural language processing (NLP) tasks. This motivates us to undertake a thorough evaluation of prominent open-source LLMs in the text-to-SQL task, aiming to gauge their viability as alternatives. However, following an assessment utilizing a standardized prompt, we observed that open-source models still exhibit a substantial performance gap compared to closed-source models. In particular, the popular open-source model CodeLLaMA-13B-Instruct demonstrates a 30\% lower execution accuracy than GPT-4 on the BIRD~\cite{bird} benchmark.

Developing specialized text-to-SQL models built upon open-source LLMs that attains performance levels comparable to close-source models holds critical importance in contexts marked by sensitivity to policy, privacy, and resource limitations.
To this end, we focus on supervised fine-tuning (SFT) to enhance the text-to-SQL capabilities of open-source base models. However, enhancing the text-to-SQL ability of open-source models through SFT remains an open challenge. A significant barrier to this progress is the high cost of achieving text-to-SQL data, which relies on manual expert annotation. The generation of high-quality text-to-SQL fine-tuning data should consider two primary perspectives. First, the inclusion of diverse data aims to facilitate cross-domain generalization, allowing the model to be successfully applied to new domains or databases. Second, alignment with executors becomes crucial to better enable the model to learn SQL from execution feedback, particularly from errors, mirroring how humans often learn from their mistakes.

In response to the data scarcity challenge, numerous endeavors~\cite{self-instruct,evol-instruct,oss-instruct} have sought to generate synthetic data utilizing larger and more powerful LLMs (strong models), such as GPT-4, creating what is denoted as \textbf{strong data}. Although strong data inherently contributes to the enhancement of data diversity~\cite{evol-instruct}, a critical factor for the domain generalization of models, its application in the text-to-SQL task remains unexplored. Additionally, the generation of valuable erroneous text-to-SQL data poses a separate challenge. Strong models often exhibit significant efforts toward correct alignment and safety, making it difficult to elicit erroneous samples. Consequently, we redirect our focus towards smaller, less well-aligned open-source models (weak models). Weak models produce valuable weak SQL samples, which can subsequently be validated and subjected to error induction with the assistance of executors. Preference learning~\cite{dpo} is employed to instruct language models to learn from both correct and incorrect samples, constituting what we refer to as \textbf{weak data}.

To verify the effectiveness of our \sense, we conduct SFT on a popular open-source base model, i.e., CodeLLaMA~\cite{codellama}, and obtain a new specialized model named \sense. We comprehensively evaluate \sense's performance on the text-to-SQL tasks, achieving state-of-the-art (SOTA) results on both the standard benchmark Spider~\cite{yu-etal-2018-spider} and the challenging benchmark BIRD~\cite{bird}, narrowing the gap between open-source and closed-source models. Additionally, we evaluate \sense on three robustness datasets: SYN~\cite{Syn}, REALISTIC~\cite{deng-etal-2021-structure}, and DK~\cite{DK}, demonstrating its advantages in robustness. Moreover, we conduct an in-depth experimental analysis offers insights into the influence of synthetic data on model performance. In summary, our contributions are threefold: 
\begin{itemize}
    \item We first evaluate both open-source and closed-source LLMs on text-to-SQL benchmarks using a standardized prompt. We observe that the text-to-SQL capabilities of open-source models were significantly inferior. It motivated us to train our specialized model \sense through SFT on an open-source LLM.
    \item We propose a synthetic data approach that uses strong models to generate strong data to enhance data diversity and employs weak models to generate weak data combined with an executor to learn from feedback.
    \item Extensive experiments shows the effectiveness of \sense, achieving SOTA performance, even competing with methods based on GPT-4. We believe that making these data and models publicly available can contribute to the advancement of the text-to-SQL community. 
\end{itemize}

\section{Preliminaries}
\paragraph{Notation Definition}

The objective of the text-to-SQL task is to convert a natural language (NL) question $Q$ into the corresponding SQL query $Y$, grounded in the database schema $\mathcal{S}=\langle\mathcal{T}, \mathcal{C}, \mathcal{V}\rangle$, which includes table names $\mathcal{T}$, column names $\mathcal{C}$, database values $\mathcal{V}$. In more challenging tasks such as BIRD~\cite{bird}, understanding NL questions or database values may require external knowledge, represented as $\mathcal{K}$. The current popular text-to-SQL task employs a cross-domain setting to evaluate a model's ability to generalize to new domains, ensuring there is no overlap between the domains of the training, development, and test sets.

\begin{table*}[htbp]
    \centering
    \small
    \begin{tabular}{c|ccccc}
        \toprule
        \textbf{Dataset} & \#Examples & \#Databases & \#Examples/\#Databases & Avg(\#Tokens) & Avg(\#\verb|JOIN|)  \\
        \midrule
        \textit{Spider}           & 7000 & 140  & 50.0      & 37.3      & 0.54\\
        \textit{Bird}             & 9428 & 69   & 136.6     & 64.0      & 1.02\\
        \midrule
        \rowcolor[RGB]{237,237,237}
        \textbf{Synthetic}        & 8216 & 503  & 16.3      & 60.3      & 1.13\\
        \bottomrule
    \end{tabular}
    \caption{Statistical information of the strong data in supervised fine-tuning stage. For synthetic data, similar databases has been merged based on semantic similarity.}
    \label{tab:data_stat}
\end{table*}

\begin{figure}[htbp]
\begin{tcblisting}{listing only, 
    halign=left,
    title=\textbf{\small Prompt template for text-to-SQL tasks.},
    colbacktitle=blue!30!white, 
    coltitle=black,
    listing options={basicstyle=\small\ttfamily,escapechar=|,language=SQL}
}
CREATE TABLE "list" (
  "LastName" TEXT,
  "FirstName" TEXT,
  "Grade" INTEGER, 
  "Classroom" INTEGER,
  PRIMARY KEY(LastName, FirstName)
);
/* 3 example rows:
SELECT * FROM list LIMIT 3;
LastName      FirstName   Grade   Classroom
CAR           MAUDE       2       101
KRISTENSEN    STORMY      6       112
VANDERWOUDE   SHERWOOD    3       107
*/
(...other tables omitted...)
-- External Knowledge: ...
-- Using valid SQLite and understanding External Knowledge, answer the following questions for the tables provided above.
|\texttt{Question: How many students are there?}|
|\\|
|\underline{\texttt{SELECT count(*) FROM list;}}|
\end{tcblisting}
\caption{Unified prompt~\cite{chang2023prompt} template for text-to-SQL tasks.}
\label{fig:prompt-inference}
\end{figure}

\begin{figure*}[t]
\small
\begin{tcblisting}{text only, 
    halign=left, 
    title= \textbf{Prompt for Synthesizing Strong Data}, 
    colbacktitle=blue!30!white, 
    coltitle=black
}
Your task is to generate one additional data point at the \texttt{\{the\_level\}} difficulty level, in alignment with the format of the two provided data points.\\
1. \textbf{Domain}: Avoid domains that have been over-represented in our repository. Do not opt for themes like Education/Universities, Healthcare/Medical, Travel/Airlines, or Entertainment/Media. \\
2. \textbf{Schema}: Post your domain selection, craft an associated set of tables. These should feature logical columns, appropriate data types, and clear relationships.\\
3. \textbf{Question Difficulty} - \texttt{\{the\_level\}}:\\
\quad- \textbf{Easy}: Simple queries focusing on a single table.\\
\quad- \textbf{Medium}: More comprehensive queries involving joins or aggregate functions across multiple tables.\\
\quad- \textbf{Hard}: Complex queries demanding deep comprehension, with answers that use multiple advanced features. \\
4. \textbf{Answer}: Formulate the SQL query that accurately addresses your question and is syntactically correct.\\
\textbf{Additional Guidelines}: \\
\quad- Venture into diverse topics or areas for your questions.\\
\quad- Ensure the SQL engages multiple tables and utilizes advanced constructs, especially for higher difficulty levels.\\
Ensure your submission only contains the Domain, Schema, Question, and Answer. Refrain from adding unrelated content or remarks.\\
\textit{(...examples and generations goes here...)}
\end{tcblisting}
\caption{Prompt for synthesizing strong data. The placeholder \texttt{the\_level} is filled on-the-fly by program, controlling the desired hardness level of the generated data point. For limited token consideration, we randomly draw two examples from Spider training set as few-shot demonstrations.}
\label{fig:prompt-synthetic}
\end{figure*}

\paragraph{Prompt Construction}

Achieving consistent results in text-to-SQL tasks with LLMs requires a standardized prompt structure to allow fair model comparisons. Following the guidelines from~\cite{chang2023prompt} and illustrated in Fig \ref{fig:prompt-inference}, our prompt comprises four elements: database schema, task instructions, optional external knowledge, and the natural language question. We employ the \texttt{CreateTable} method for detailing database schemas and \texttt{SelectRow} to showcase table contents, ensuring SQL keywords and schema are presented uniformly. Task instructions are clear: ``\textit{-- Using valid SQLite, answer the following questions for the tables provided above}.'' External knowledge, if necessary, precedes the question. This standardized prompt, denoted as $X$, is designed to serve as the input to the LLM.

\section{Methodology}

Overall, our approach is divided into two distinct phases. Initially, we enhance the base model's text-to-SQL capabilities through \textit{Supervised Fine-tuning (SFT)}, with a primary focus on the diversity and quality of data. We refer to this portion of data as \textbf{strong data}. Subsequently, we employ \textit{Preference Learning} to inspire the model to learn from incorrect SQLs, which we denote them as \textbf{weak data}, necessitating the use of weaker language models for error generation. The details of these two processes are described below.

\subsection{Strong Data: Supervised Fine-tuning}

Supervised Fine-tuning (SFT) will significantly enhance the model's performance in generating appropriate responses, including text-to-SQL. The currently popular cross-domain datasets, primarily Spider and BIRD, incur high annotation costs due to the need for human experts.
To mitigate this and further expand the scale, we turn to the powerful language model GPT-4 for assistance, utilizing prompts to synthesize target data. Given that cross-domain generalization is a central challenge for text-to-SQL, we designed hints to encourage diversity by prompting GPT-4 to generate sufficiently diverse datasets, as shown in Figure~\ref{fig:prompt-synthetic}. 
As illustrated in Table~\ref{tab:data_stat}, the ratio of examples per domain in our synthetic dataset is markedly lower than that observed in Spider and Bird, signifying a higher domain diversity. Additionally, the synthetic data features a higher average number of JOIN operations in SQL queries, indicating a greater complexity and depth in the constructed SQLs.
These include mechanisms for controlling question difficulty, promoting domain diversity, and explicitly excluding over-represented domains, thereby guiding GPT-4 to generate data points that are not only diverse but also tailored to various levels of complexity.

Given a stong data set $\mathcal{D}_s$ of input prompt $\boldsymbol{x}$ and target response $y$ generated, the supervised fine-tuning could be formulated as the log likelihood loss:
\begin{equation}
\mathbb{E}_{(\boldsymbol{x}, \boldsymbol{y}) \sim \mathcal{D}_s}\left[\sum_{t=1}^T \log p_\theta\left(y_t \mid \boldsymbol{y}_{1: t-1}, \boldsymbol{x}\right)\right],
\end{equation}
where $\theta$ is the parameters of language model, and $p_\theta(\boldsymbol{y} \mid \boldsymbol{x})=\prod_{t=1}^T p_\theta\left(y_t \mid \boldsymbol{y}_{<t}, \boldsymbol{x}\right)$ is the conditional probability distribution of target SQL sequence $y$ given prompt $x$. $T$ is the sequence length of $y$, and $t$ is the auto-aggressive decoding step.

\subsection{Weak Data: Preference Learning}
The second phase involves a more nuanced approach for weak data. Here, we introduce the model to incorrect SQL queries intentionally generated by weaker LLMs. Through Preference Learning~\cite{dpo}, the model is encouraged to discern between correct and incorrect SQL, effectively learning from its mistakes. This process not only refines the model's understanding of SQL syntax but also enhances its resilience to common errors that might occur in real-world scenarios.

Given a natural language description $x$, we generate an output $y'$ using weaker models (smaller in size and less well-aligned). We then execute $y'$ using an SQL executor $\textsf{E}$, and if the execution result matches the ground truth $y$, we consider it a positive sample $y_w$. Conversely, if the result is inconsistent, we label it as a negative sample $y_l$.
\begin{equation}
\begin{cases}
y_w = y', & \text{if } \textsf{E}(y') = \textsf{E}(y) \\
y_l = y', & \text{if } \textsf{E}(y') \neq \textsf{E}(y).
\end{cases}
\end{equation}

We construct a dataset $\mathcal{D}_w$ with both positive and negative examples and optimize the model using the recently popular preference learning method, direct preference optimization~\cite[DPO,][]{dpo}. DPO fine-tunes the model directly based on preference data, bypassing the reward modeling stage and aiming to maximize the following objective function:
\begin{equation}
\scalebox{0.848}{$
\mathop{\mathbb{E}}\limits_{(x, y_w, y_l) \sim \mathcal{D}_w} \log \sigma\left(\beta \log \frac{p_\theta(y_w \mid x)}{p_{\mathrm{ref}}(y_w \mid x)} - \beta \log \frac{p_\theta(y_l \mid x)}{p_{\mathrm{ref}}(y_l \mid x)}\right)
$}
\end{equation}
where $p_\theta$ represents the probability distribution of the target model's predictions, $p_{\mathrm{ref}}$ denotes the probability distribution from a reference model, and $\beta$ is a parameter that regulates the extent of the target model's divergence from the reference model.
By training on the preference dataset, we can align LLM with executor preferences. 

Leveraging the described methodology, we conducted two phases of training using CodeLLaMA-7B and CodeLLaMA-13B, successfully yielding \sense{-7B} and \sense{-13B} as final models.

\section{Experiments}

\newcommand{\bv}[1]{\textit{\textbf{#1}}}
\newcommand{\compname}[1]{\multicolumn{2}{l|}{#1}}
\newcommand{\addone}[1]{\multicolumn{2}{l|}{{\quad{#1}}}}
\newcommand{\addtwo}[2]{\multicolumn{1}{l}{\quad#1} & \multicolumn{1}{l|}{#2}}
\newcommand{\weakdpsk}{\bv{Weak-1.3B}}
\newcommand{\weakqwen}{\setlength\fboxsep{1pt}\colorbox{red!30}{\bv{Weak$^{\textrm{Q}}$-1.8B}}}
\newcommand{\headercell}{\multicolumn{2}{c|}{\multirow{2}{*}{\bv{Model / Method}}}}
\newcommand{\cellaligned}[2]{\multicolumn{1}{#1}{#2}}

\newcommand{\rtt}[1]{-\texttt{#1}}

\begin{table*}[!h]
    \centering
    \small
    \begin{tabular}{l|x{1.2cm}x{1.2cm}x{1.2cm}|x{1.2cm}x{1.2cm}}
    \toprule
    \multirow{2}{*}{\quad\vspace{-2mm}\bv{Model / Method}} 
    & \multicolumn{3}{c|}{\bv{Spider}} & \multicolumn{2}{c}{\bv{Bird}} \\
    \cmidrule{2-6}
    & \bv{Dev-EX} & \bv{Dev-TS} & \bv{Test} & \bv{Dev} & \bv{Test} \\
    \midrule
    \multicolumn{6}{c}{\bv{Prompting Methods w/ Closed-Source LLMs}} \\
    \midrule
    PaLM-2~\cite{palm2}                 &  -   &  -   & -    & 27.4 & 33.1 \\ 
    Claude-2~\cite{claude2}               &  -   &  -   & -    & 42.7 & 49.0 \\
    ChatGPT~\cite{chatgpt}                & 72.3 &  -   & -    & 36.6 & 40.1 \\
    GPT-4~\cite{gpt4}                  & 72.9 & 64.9 & -    & 49.2 & 54.9 \\
    Few-shot SQL-PaLM~\cite{sqlpalm} & 82.7 & 77.3 & -    & -    & -    \\
    DIN-SQL + GPT-4~\cite{dinsql}          & 82.8 & 74.2 & \underline{85.3} & 50.7 & 55.9 \\
    ACT-SQL + GPT-4~\cite{actsql}          & 82.9 & 74.5 & -    & -    & -    \\
    DAIL-SQL + GPT-4~\cite{dailsql}         & \underline{83.5} &  76.2  & \textbf{86.6} & \underline{54.8} & 57.4 \\
    \midrule
    \multicolumn{6}{c}{\bv{Fine-tuning Models}} \\
    \midrule
    T5-3B + PICARD~\cite{PICARD}        & 79.3 & 69.4 & 75.1 & - & - \\
    RASAT + PICARD~\cite{rasat}             & 80.5 & 70.3 & 75.5 & - & - \\
    RESDSQL-3B + NatSQL~\cite{resdsql}   & \textbf{84.1} & 73.5 & 79.9 & - & - \\
    Fine-tuned SQL-PaLM~\cite{sqlpalm} & 82.8 & 78.2 & - & - & - \\
    \midrule
    \multicolumn{6}{c}{\bv{Open-Source LLMs}} \\
    \midrule
    LLaMA2-7B~\cite{llama2}                  & 28.0 & 23.8 & - &  7.1 & - \\
    LLaMA2-7B-Chat~\cite{llama2}             & 36.9 & 34.9 & - & 11.3 & - \\
    Qwen-1.8B~\cite{bai2023qwen}                  & 54.8 & 48.6 & - & 13.2 & - \\
    LLaMA2-13B-Chat~\cite{llama2}            & 49.6 & 45.5 & - & 14.2 & - \\
    LLaMA2-13B~\cite{llama2}                 & 47.4 & 39.4 & - & 15.3 & - \\
    StarCoder-3B~\cite{starcoder}               & 52.7 & 47.0 & - & 15.3 & - \\
    StarCoder-7B~\cite{starcoder}               & 60.7 & 55.1 & - & 17.2 & - \\
    DeepSeek-Coder-1.3B~\cite{deepseekcoder}        & 59.3 & 53.2 & - & 22.0 & - \\
    $^\clubsuit$CodeLLaMA-7B~\cite{codellama}   & 61.1 & 52.3 & - & 22.5 & - \\
    $^\heartsuit$CodeLLaMA-13B~\cite{codellama} & 61.7 & 53.5 & - & 22.9 & - \\
    CodeLLaMA-7B-Instruct~\cite{codellama}      & 63.4 & 54.2 & - & 23.0 & - \\
    DeepSeek-Coder-1.3B-Instruct~\cite{deepseekcoder}& 53.2 & 48.7 & - & 24.1 & - \\
    StarCoder-15B~\cite{starcoder}              & 63.9 & 57.9 & - & 24.4 & - \\
    CodeLLaMA-13B-Instruct~\cite{codellama}     & 62.3 & 52.5 & - & 24.7 & - \\
    Qwen-7B~\cite{bai2023qwen}                    & 63.6 & 54.5 & - & 26.1 & - \\
    \midrule
    \multicolumn{6}{c}{\bv{Ours}} \\
    \midrule
    \rowcolor[RGB]{237,237,237}
    $^\clubsuit$\sense{-7B}   & 83.2 & \underline{81.7} & 83.5 & 51.8 & \underline{59.3} \\
    \rowcolor[RGB]{237,237,237}
    $^\heartsuit$\sense{-13B} & \textbf{84.1} & \textbf{83.5} & \textbf{86.6} & \textbf{55.5} & \textbf{63.4} \\
    \bottomrule
    \end{tabular}
    \caption{Performance comparison on Spider and Bird benchmarks. To show the relationship of our final presented models and base models, we denote them by $^\clubsuit$ and $^\heartsuit$. Specifically, \sense{-7B} is based on CodeLLaMA-7B and \sense{-13B} is based on CodeLLaMA-13B.}
    \label{tab:result_overall}
\end{table*}

\subsection{Evaluation Benchmarks}

We evaluated the effectiveness of \sense using popular text-to-SQL benchmarks across five datasets.
\paragraph{General Benchmark} Spider~\cite{yu-etal-2018-spider} comprises 7,000 Text-SQL pairs in its training set and 1,034 pairs in its development set, across 200 different databases and 138 domains. 
\paragraph{Challenge Benchmark} BIRD~\cite{bird} is a new benchmark of large real-world databases, containing 95 large databases with high-quality Text-SQL pairs, totaling 33.4GB of data across 37 fields. Unlike Spider, BIRD focuses on massive and real database contents, the external knowledge reasoning between natural language questions and database content.
\paragraph{Robust Benchmarks} SYN~\cite{Syn}  replaces simple string-matched problem tags or pattern names with their synonyms. DK~\cite{DK} requires the text-to-SQL parser to have domain knowledge reasoning capabilities. REALISTIC~\cite{deng-etal-2021-structure} replaces mentioned schema items in questions to make them closer to real-world scenarios.

\begin{table*}
    \centering
    \small
    \begin{tabular}{l|x{1.5cm}x{1.5cm}|x{1.5cm}x{1.5cm}|x{1.5cm}|c}
    \toprule
    \multirow{2}{*}{\quad\vspace{-2mm}\bv{Model / Method}}  & \multicolumn{2}{c|}{\bv{SYN}} & \multicolumn{2}{c|}{\bv{REALISTIC}} & \bv{DK} & \multirow{2}{*}{\vspace{-2mm}\bv{Average}} \\  
    \cmidrule{2-6}
    & \bv{EX} & \bv{TS} & \bv{EX} & \bv{TS} & \bv{EX} & \\
    \midrule
    RESDSQL-3B+NatSQL   & \underline{76.9} & 66.8 & 81.9 & 70.1 & 66.0 & 72.3 \\
    Few-shot SQL-PaLM   & 74.6 & \underline{67.4} & 77.6 & 72.4 & 66.5 & 71.7 \\
    Fine-tuned SQL-PaLM & 70.9 & 66.4 & 77.4 & 73.2 & 67.5 & 71.1 \\
    \midrule
    \rowcolor[RGB]{237,237,237}
    \sense{-7B}               & 72.6 & 64.9 & \underline{82.7} & \underline{75.6} & \underline{77.9} & \underline{74.7} \\
    \rowcolor[RGB]{237,237,237}
    \sense{-13B}              & \textbf{77.6} & \textbf{70.2} & \textbf{84.1} & \textbf{76.6} & \textbf{80.2} & \textbf{77.7} \\
    \bottomrule
    \end{tabular}
    \caption{Evaluation of \sense and various previously proposed methods on Spider-based robustness benchmarks: Spider-SYN, REALISTIC and Spider-DK. TS is not reported for DK due to incompatibility.}
    \label{tab:result_robustness}
\end{table*}

\newcommand{\wo}{\quad\textit{w/o}~}

\begin{table*}[htbp]
    \centering
    \small
    \begin{tabular}{l|x{1.5cm}x{1.5cm}x{1.5cm}c|x{1.5cm}}
        \toprule
        \bv{Model / Method} & \bv{Easy} & \bv{Medium} & \bv{Hard} & \bv{Extra Hard} & \bv{All} \\
        \midrule
        DIN-SQL + GPT-4          & 91.1 & 79.8 & 64.9 & 43.4 & 74.2 \\
        ACT-SQL + GPT-4          & 91.1 & 79.4 & 67.2 & 44.0 & 74.5 \\
        DAIL-SQL + GPT-4         & 90.3 & 81.8 & 66.1 & \underline{50.6} & 76.2 \\
        Few-shot SQL-PaLM   & \underline{93.5} & 84.8 & 62.6 & 48.2 & 77.3 \\
        Fine-tuned SQL-PaLM & \underline{93.5} & \underline{85.2} & \underline{68.4} & 47.0 & \underline{78.2} \\
        \midrule
        \rowcolor[RGB]{237,237,237}
        \sense{-13B}              & \textbf{95.2} & \textbf{88.6} & \textbf{75.9} & \textbf{60.3} & \textbf{83.5} \\
        \bottomrule
    \end{tabular}
    \caption{Test Suite accuracy on Spider Dev, categorized by SQL hardness levels.}
    \label{tab:hardness_spider}
\end{table*}

\subsection{Evaluation Metrics}

For Spider and its robustness benchmarks, we follow Spider's official evaluation protocol \footnote{\url{https://yale-lily.github.io/spider}}, using EX~\cite{yu-etal-2018-spider} and TS~\cite{spiderts} metrics\footnote{TS is not reported for DK due to incompatibility.}. EX measures if the SQL output exactly matches the execution result of provided golden SQL. TS is a more reliable metric that confirms if a SQL query passes all EX checks on various tests created through database augmentation. For BIRD, we adopt its official evaluation scripts\footnote{\url{https://bird-bench.github.io}}, focusing on EX accuracy evaluation.

\subsection{Compared Methods}

We compared a variety of baseline methods, which can be categorized into three categories. 
\paragraph{Prompting Methods} 
ACT-SQL~\cite{actsql} introduced a method for automatically generating Chain of Thought (CoT)~\cite{cot} examples. DIN-SQL~\cite{dinsql} uses prompts to break down the complex text-to-SQL task into smaller subtasks for enhanced performance. DAIL-SQL~\cite{dailsql} made improvements in question representation, example selection, and sample sequence organization.
\paragraph{Fine-tuning Models} 
PICARD~\cite{PICARD} is a constrained decoding approach fine-tuned on T5-3B. RASAT~\cite{rasat} and Graphix~\cite{graphixt5} focus on how to incorporate structure information into the fine-tuning process of the T5 model~\cite{t5-jmir-2020}, while RESDSQL-3B~\cite{resdsql} decouples schema linking and skeleton parsing.
\paragraph{Open-Source LLMs} 
Recently, there has been a surge in open-source LLMs. We selected some of the most recently popular LLMs, including various sizes and versions of DeepSeek-Coder~\cite{deepseekcoder}, Qwen~\cite{bai2023qwen}, StarCoder~\cite{starcoder}, LLaMA2~\cite{llama2}, and CodeLLaMA~\cite{codellama}. We adopted a unified prompt as shown in Figure~\ref{fig:prompt-inference} to ensure a fair comparison with \sense.

\begin{figure*}[htb]
    \centering
    \includegraphics[width=1.0\textwidth]{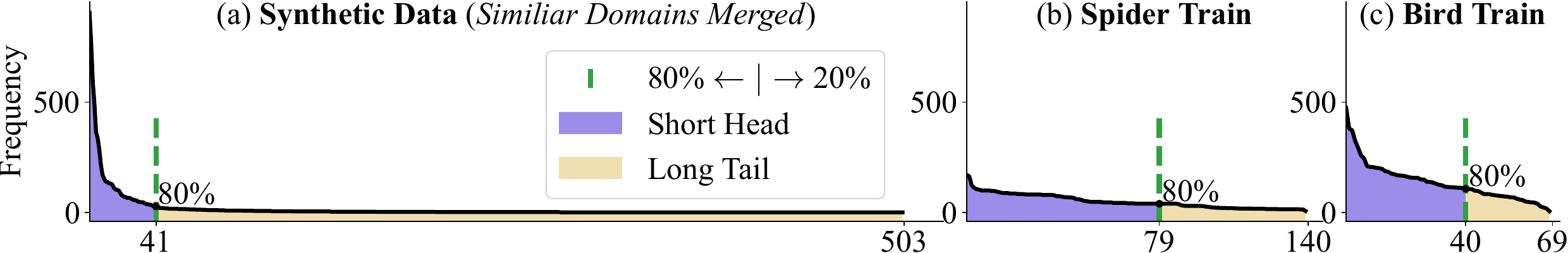}
    \caption{Domain density comparison. This visualization sorts domains by example count, showcasing a long-tail distribution to highlight the broad diversity within our synthetic dataset.
    }
    \label{fig:long-tail}
\end{figure*}

\begin{figure}[htb]
    \centering
    \includegraphics[width=0.75\linewidth]{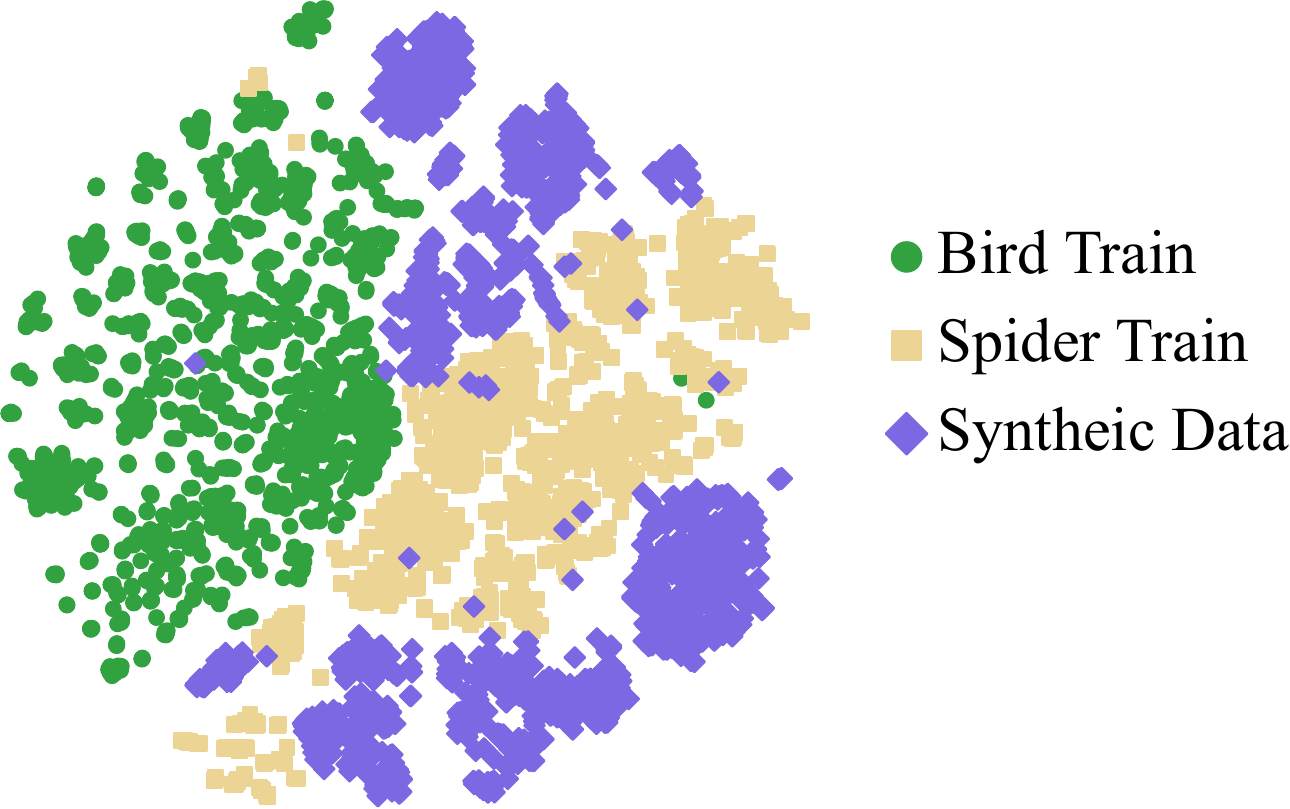}
    \caption{2-D t-SNE visualization comparing original and synthetic data's last-layer hidden representations post-supervised fine-tuning on last token.}
    \label{fig:tsne}
\end{figure}

\begin{figure}[htb]
    \includegraphics[width=1.0\linewidth]{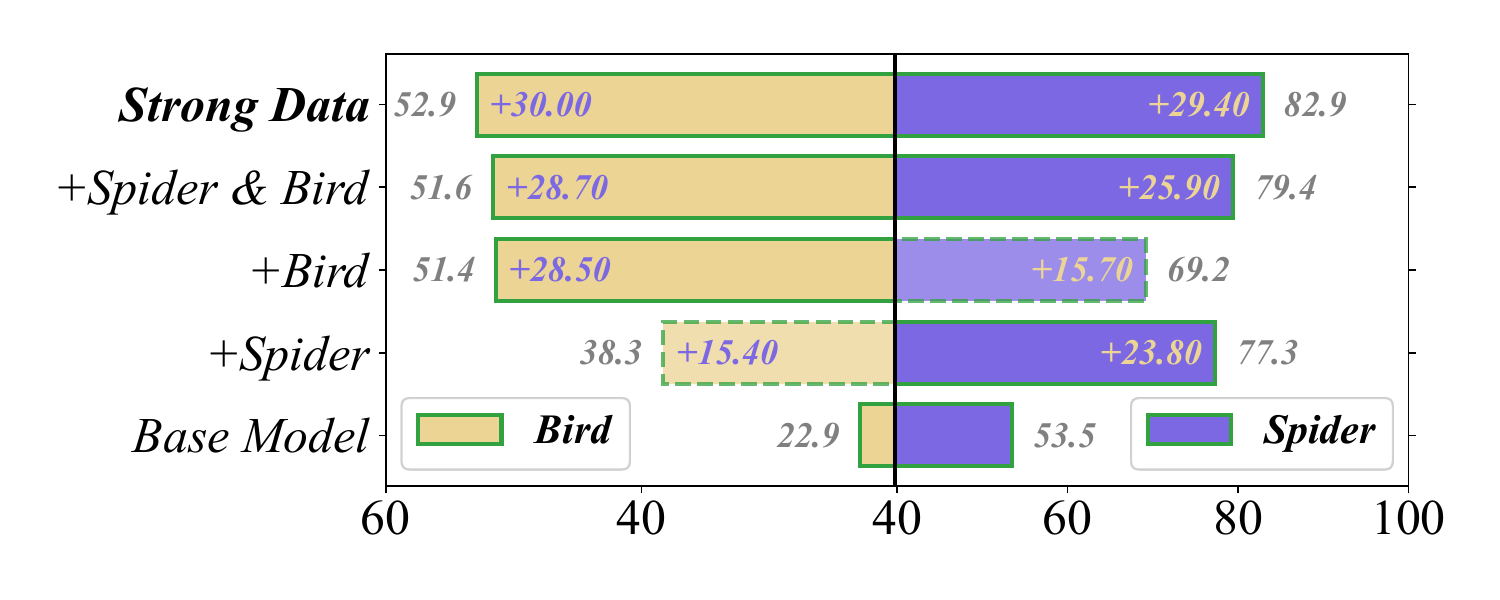}
    \caption{Bird dev and Spider dev scores for different Supervised fine-tuning data on CodeLLaMA-13B. We report EX and TS for Bird and Spider, respectively.}
    \label{fig:ablation-bar}
\end{figure}

\subsection{Implementation Details}

We choose CodeLLaMA-7B and CodeLLaMA-13B as our primary models, and DeepSeek-Coder-1.3B as a weak model to generate preference data. Our experiments were run on 8$\times$A100 GPUs, combining datasets from Spider and Bird with GPT-4-generated data for supervised fine-tuning using the AdamW optimizer with a learning rate of 2e-5 and a cosine warmup scheduler over three epochs.
The preference learning phase starts with the generation of the weak data through the fine-tuned weak model and a SQL evaluator. The evaluator recognize each generated SQL as positive or negative, thus further enabled us to craft a preference dataset. This dataset served as the foundation for Direct Preference Optimization~\cite[DPO,][]{dpo} training. For DPO training, the Adam optimizer was selected, with the learning rate adjusted to 2e-6 and the $\beta$ parameter maintained at 0.2, consistent with the original DPO settings.

\subsection{Overall Performance}

\paragraph{Results on General Settings}

Table~\ref{tab:result_overall} shows prompting methods surpass fine-tuning in text-to-SQL, thanks to closed-source LLMs and tailored prompts. Open-source LLMs lag in generalization. Larger models correlate with better outcomes, and instruction tuning boosts performance, showcasing synthetic data's tuning utility. Remarkably, \sense achieves state-of-the-art (SOTA) results on the Spider dataset, surpassing the GPT-4-based DAIL-SQL. Specifically, \sense{-13B} shows a 21.8\% improvement over CodeLLaMA-13B-Instruct in the development set and marginally exceeds DAIL-SQL, indicating \sense's potential in narrowing the performance gap between open and closed-source models in text-to-SQL challenges.

\paragraph{Results on Challenge Settings}
Experiments on BIRD reveal its complexity, none of the open-source LLMs perform well on BIRD, yet \sense{-13B} sets a new standard, outperforming DAIL-SQL by 5.98\% on the test set, as Table~\ref{tab:result_overall} shows. This highlights the benefits of specialized open-source LLMs for challenging environments.

\paragraph{Results on Robust Settings}
Table~\ref{tab:result_robustness} reveals \sense excels in robustness (SYN, DK, REALISTIC) even without extra training. \sense{-7B} and \sense{-13B} lead, surpassing RESDSQL-3B by 1.4\% and 5.4\% on average. Notably, \sense's strength in DK suggests synthetic data effectively leverages the base model's domain knowledge.

\subsection{Fine-grained Analysis on Hardness}
Spider's difficulty labels reveal \sense{-13B}'s superiority across all levels, as shown in Table~\ref{tab:hardness_spider}. Performance gains over the best alternatives are notable: Easy (1.6\%), Medium (3.4\%), Hard (7.5\%), and Extra Hard (9.7\%). This suggests that it has an advantage in processing hard samples, benefiting from the difficulty control in synthetic data prompt.

\subsection{Ablation Study}

Table~\ref{tab:result_dev} presents an ablation study on \sense, dissecting its components to assess their individual impact. The study focuses on three key topics.

\begin{table}[htb]
    \centering
    \small
    \begin{tabular}{l|cc|c}
    \toprule
    \multirow{2}{*}{\bv{Model}}  & \multicolumn{2}{c|}{\bv{Spider-Dev}} & \cellaligned{c}{\bv{Bird}} \\
    & \cellaligned{c}{\bv{EX}} & \cellaligned{c|}{\bv{TS}} & \cellaligned{c}{\bv{Dev}} \\ 
    \midrule
    \multicolumn{4}{c}{\itshape \textcolor{gray}{Ablation for Main Results}} \\
    \midrule
    DeepSeek-Coder-1.3B$^\clubsuit$ & 79.1 & 80.2 & 40.7 \\
    \wo\textbf{Strong}                   & 59.3 & 53.2 & 22.0 \\
    \midrule
    \rowcolor[RGB]{237,237,237}
    \sense{-7B}         & 83.2 & 81.7 & 51.8 \\
    \wo\textbf{Weak}   & 82.3 & 81.6 & 49.9 \\
    \wo\textbf{Strong} & 61.1 & 52.3 & 22.5 \\
    \midrule
    \rowcolor[RGB]{237,237,237}
    \sense{-13B}        & 84.0 & 83.5 & 55.5 \\
    \wo\textbf{Weak}   & 83.9 & 82.9 & 52.9 \\
    \wo\textbf{Strong} & 61.7 & 53.5 & 22.9 \\
    \midrule
    \multicolumn{4}{c}{\itshape \textcolor{gray}{Transferability on Homogeneous Models}} \\
    \midrule
    Qwen-1.8B$^\clubsuit$ & 76.2 & 75.2 & 38.1 \\
    \wo\textbf{Strong}         & 54.8 & 48.6 & 13.2 \\
    \midrule
    \rowcolor[RGB]{237,237,237}
    \senseq{-7B}                & 84.2 & 84.6 & 52.5 \\
    \wo\textbf{Weak}           & 83.9 & 83.4 & 50.5 \\
    \wo\textbf{Strong}         & 63.1 & 54.4 & 26.1 \\
    \bottomrule
    \end{tabular}
    \caption{Ablation and transferability results. Upper section details the effects of excluding \textbf{weak} and \textbf{strong data} in fine-tuning. Lower section assesses model transferability, using Qwen-1.8B and Qwen-7B. $^\clubsuit$ marks a model fine-tuned with strong data.}
    \label{tab:result_dev}
\end{table}

\begin{figure*}[!h]
    \centering
    \includegraphics[width=0.93\linewidth]{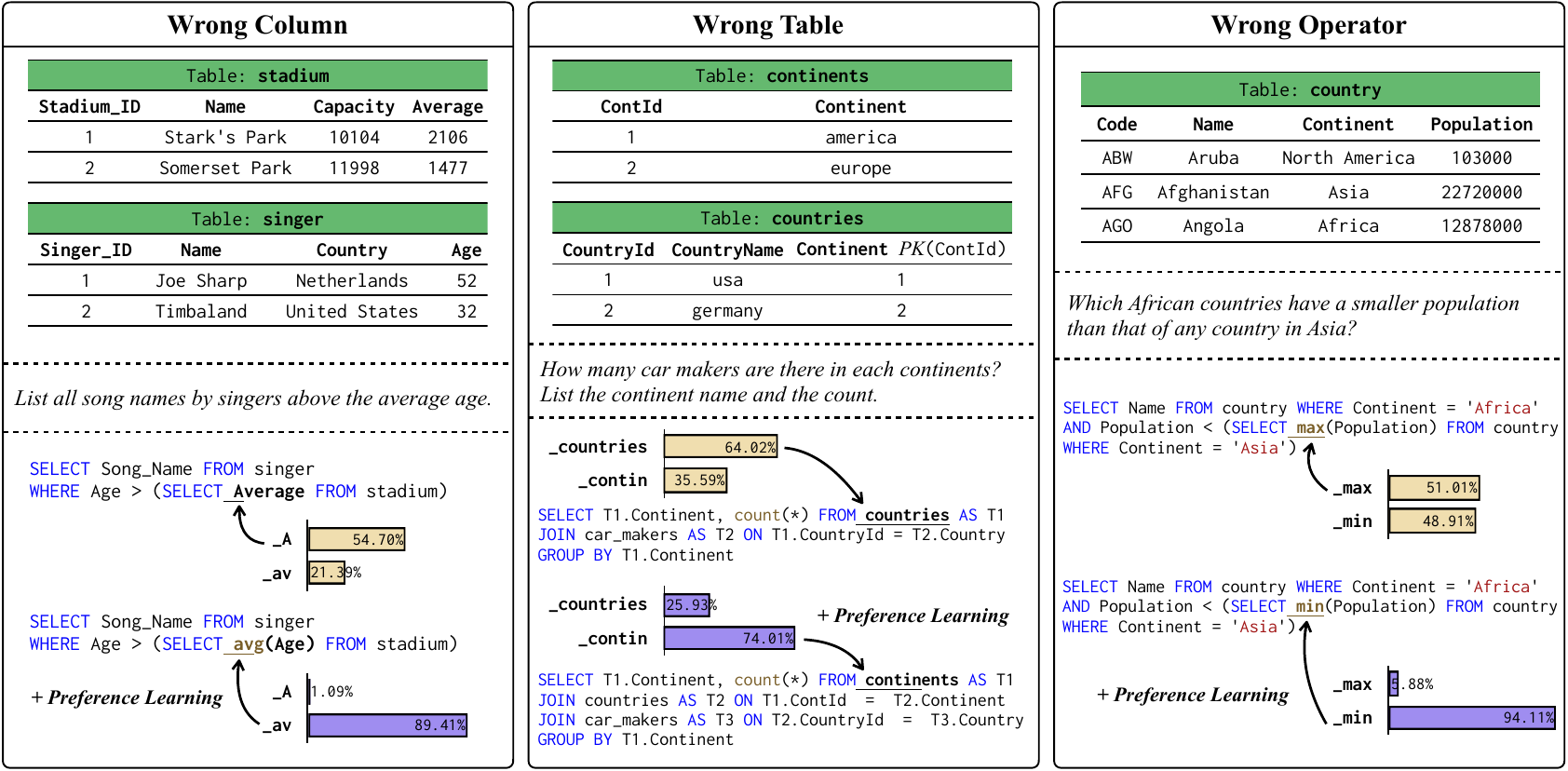}
    \caption{Preference learning enhances text-to-SQL performance in critical tokens.}
    \label{fig:case-study}
\end{figure*}

\begin{table*}[htbp]
    \centering
    \small
    \scalebox{1}{
    \begin{tabular}{l|cccc|c}
        \toprule
        \bv{Model} 
        & \makecell{\bv{MMLU}\\5-shot} 
        & \makecell{\bv{ARC-Challenge}\\25-shot} 
        & \makecell{\bv{GSM8K}\\8-shot, CoT} 
        & \makecell{\bv{HumanEval}\\Greedy}
        & \bv{Average} \\
        \midrule
        CodeLLaMA-7B & 39.2 & 42.1 & 11.4 & 29.3 & 30.5 \\
        \rowcolor[RGB]{237,237,237}
        \sense{-7B} & 39.1 & 42.5 & 11.0 & 31.7 & 31.1 \\
        \midrule
        CodeLLaMA-13B & 43.7 & 46.4 & 21.8 & 34.2 & 36.5 \\
        \rowcolor[RGB]{237,237,237}
        \sense{-13B} & 42.7 & 47.9 & 18.1 & 40.2 & 37.2 \\
        \bottomrule
    \end{tabular}}
    \caption{Performance of \sense and corresponding base models on MMLU, ARC-Challenge, GSM8K and HumanEval.}
    \label{tab:board_tasks}
\end{table*}

\paragraph{Why Strong Data is Helpful?}

Analysis from Figure~\ref{fig:ablation-bar} and Table~\ref{tab:result_dev} shows strong data significantly boosts Spider accuracy due to its simpler SQL queries and the emphasis on domain generalization, and it can be seen that the data of bird and spider can mutually enhance each other when the data in the corresponding field is absent. Figure~\ref{fig:long-tail} illustrates strong data's broader long-tail distribution, enabled by LLMs' stored knowledge, enhancing \sense's ability to adapt to new domains. Additionally, t-SNE visualizations in Figure~\ref{fig:tsne} highlight synthetic samples' role in bridging the gaps left by human-annotated data, further validating the value of synthetic data.

\paragraph{Why Weak Data is Helpful?}
Weak data, when used with preference learning, helps \sense to refine its output by learning from errors, ensuring closer alignment with the SQL executor. As indicated in Table~\ref{tab:result_dev}, weak data significantly enhances overall performance, particularly by improving the complexity of SQL queries generated, with a notable 4.9\% uplift on BIRD when compared to strong data-trained models. Further, Figure~\ref{fig:case-study} and case studies reveal weak data's role in reducing hallucinations in SQL generation, minimizing errors in selecting columns, tables, and operators, vital for crafting intricate SQL commands.

\paragraph{Transferablity across Different LLMs}

In Table~\ref{tab:result_overall}, \sense is initialized with CodeLLaMA and utilizes the smaller DeepSeek-Coder as the generator for weak data. From the perspective of model differences, CodeLLaMA and DeepSeek-Coder can be considered heterogeneous models due to their distinct structural details and pre-training data. This make our curiosity about whether \sense could effectively transfer to homogeneous models with identical pre-training data. We selected the Qwen series, a family of open-source models with a rich variety of sizes. We used Qwen-7B as the base model and Qwen-1.8B as the weak model for synthesizing, creating a new variant, \senseq. As shown in Table~\ref{tab:result_dev}, we found that the approach of using synthetic data works equally well under homogeneous models, demonstrating the same level of improvement as \sense. This confirms the transferability of the proposed method.

\paragraph{Performance on General and Board Tasks} 

In addition to evaluating the performance of \sense on several text-to-SQL tasks, we assessed its generalization capabilities through experiments on several benchmarks: MMLU~\cite{hendryckstest2021} for language understanding, ARC-Challenge~\cite{clark2018think} for common sense reasoning, GSM8K~\cite{cobbe2021training} for math reasoning, and HumanEval~\cite{chen2021evaluating} for code generation.
The results, presented in Table~\ref{tab:board_tasks}, highlight \sense's competitive performance across these diverse tasks, demonstrating its broad applicability beyond the SQL domain. Notably, while \sense maintaining competitive performance on math reasoning, the \sense{-13B} exhibits significant improvement in code generation performance compared to its base models. These findings underscore \sense's robust versatility and generalization capabilities across various tasks. Furthermore, our experiments show that \sense models maintain performance on MMLU tasks, even when specifically fine-tuned for text-to-SQL tasks, confirming our proposed method won't affect the knowledge stored in the LLMs. Moreover, if there is a need to enhance NLP task performance, additional general data could be incorporated through instruction fine-tuning, though this is beyond the scope of this paper.

\section{Related Work}

\paragraph{Text-to-SQL Parsing}

In the realm of text-to-SQL parsing, early methods like IRNET~\cite{irnet} focused on learning representations using attention-based models, while subsequent works introduced fine-tuning based models~\cite{GNNSQL,rat-sql,lgesql,s2sql,resdsql,liu-etal-2021-awakening,shi2021learning}.
Other advancements~\cite{shaw-etal-2021-compositional, PICARD,UnifiedSKG} have leveraged T5, demonstrating their effectiveness.
Recently, the emergence of LLMs\cite{chatgpt,gpt4,palm2,claude2} has garnered significant attention. Building on top of these models, various works have explored innovative prompting techniques. Such as the automatic generation of Chain-of-Thought~\cite{cot} examples by ACT-SQL~\cite{actsql}, the decomposition of complex tasks into sub-tasks by DIN-SQL~\cite{dinsql}, and sample organization by DAIL-SQL~\cite{dailsql} have significantly improved performance in the text-to-SQL domain. TAP4LLM~\cite{sui2023tap4llm} proposed a table provider to better perform semi-structured data reasoning. \cite{tai-etal-2023-exploring} study how to enhance reasoning ability CoT style prompting. Our work sets apart by leveraging open-source LLMs, matching the prowess of proprietary models in text-to-SQL tasks.

\paragraph{Synthetic Data}
There are many methods have investigated the usage of LLMs to synthesize data. Self-Instruct~\cite{self-instruct} introduced a framework for improving the instruction-following capabilities. \citet{yue2023mammoth} managed to use hybrid rationales in developing advanced math models. \citet{yu2024metamath} generated plenty of mathematical questions by rewriting from multiple aspects. \citet{yuan2024scaling} uses supervised models to bootstrap more augmented samples for math. Our approach, distinct from others, utilizes both large and smaller models for data generation, showcasing efficacy in text-to-SQL tasks.

\section{Conclusion}
In this paper, we proposed a novel model \sense to explore synthetic data in text-to-SQL parsing. By combining synthetic strong data from larger models with weak data from smaller models, \sense enhances domain generalization and learns from executor feedback through preference learning. Extensive experiments demonstrate that \sense achieves state-of-the-art performance on well-known benchmarks, significantly narrowing the gap between open-source models and closed-source models. The release of \sense data and models aims to further the progress in the text-to-SQL domain, highlighting the potential of open-source LLM to be fine-tuned with synthetic data.

\section*{Limitations} 
Although our method has shown promising results and significant progress in various aspects, it's crucial to explore the potential limitations. Firstly, due to limited computational resources and time constraints, we were unable to fine-tune our method on larger language models, such as LLaMA2-70B. The effectiveness of synthesizing data on larger models remains unclear. Secondly, our evaluation mainly focused on the text-to-SQL task. However, the potential of our data synthesis technique across diverse tasks, including code generation and math problems, which also benefit from execution-based validation, remains to be fully examined.

\section*{Acknowledgments}
Min Yang was supported by National Key Research and Development Program of China (2022YFF0902100), National Natural Science Foundation of China (62376262), the Natural Science Foundation of Guangdong Province of China (2024A1515030166), Shenzhen Science and Technology Innovation Program (KQTD20190929172835662), Shenzhen Basic Research Foundation (JCYJ20210324115614039).
This work was supported by Alibaba Group through Alibaba Research Intern Program.
This work was supported by Alibaba Group through Alibaba Research Intern Program.

\clearpage
\bibliography{custom}

\end{document}